\documentclass{article}

\usepackage{microtype}
\usepackage{graphicx}
\usepackage{subfigure}
\usepackage{booktabs} 
\usepackage{bm}

\usepackage{hyperref}
\newcommand{\ssdiff}{\textsc{CPDiffusion-SS}}
\newcommand{\ie}{\textit{i.e.}, }
\def\gG{{\mathcal{G}}}
\def\gP{{\mathcal{P}}}
\def\mA{{\bm{A}}}

\def\mE{{\bm{E}}}

\def\mH{{\bm{H}}}

\def\mM{{\bm{M}}}

\def\mS{{\bm{S}}}

\def\mX{{\bm{X}}}

\def\mZ{{\bm{Z}}}

\def\va{{\bm{a}}}

\def\ve{{\bm{e}}}

\def\vh{{\bm{h}}}

\def\vm{{\bm{m}}}

\def\vs{{\bm{s}}}

\def\vx{{\bm{x}}}

\def\vz{{\bm{z}}}

\def\rva{{\mathbf{a}}}

\def\rmA{{\mathbf{A}}}

\def\sP{{\mathbb{P}}}

\newcommand{\softmax}{\mathrm{softmax}}

\newcommand{\Ls}{\mathcal{L}}
\newcommand{\KL}{D_{\mathrm{KL}}}

\usepackage[accepted]{icml2024}

\usepackage{amsmath}
\usepackage{amssymb}
\usepackage{mathtools}
\usepackage{amsthm}

\usepackage[capitalize,noabbrev]{cleveref}

\theoremstyle{plain}

\theoremstyle{definition}

\theoremstyle{remark}

\usepackage[textsize=tiny]{todonotes}

\icmltitlerunning{Secondary Structure-Guided Novel Protein Sequence Generation with Latent Graph Diffusion}

\begin{document}

\twocolumn[
\icmltitle{Secondary Structure-Guided Novel Protein Sequence Generation with Latent Graph Diffusion}



\icmlsetsymbol{equal}{*}

\begin{icmlauthorlist}
\icmlauthor{{Yutong} {Hu}}{equal,2}
\icmlauthor{{Yang} {Tan}}{equal,1,3,4}
\icmlauthor{{Andi} {Han}}{equal,9}
\icmlauthor{{Lirong} {Zheng}}{6}
\icmlauthor{{Liang} {Hong}}{2,1,4}
\icmlauthor{{Bingxin} {Zhou}}{2,1}
\end{icmlauthorlist}

\icmlaffiliation{1}{Shanghai National Center for Applied Mathematics (SJTU center)}
\icmlaffiliation{2}{Shanghai Jiao Tong University}
\icmlaffiliation{3}{East China University of Science and Technology}
\icmlaffiliation{4}{Shanghai Artificial Intelligence Laboratory}
\icmlaffiliation{6}{University of Michigan Medical School}
\icmlaffiliation{9}{RIKEN AIP}

\icmlcorrespondingauthor{Bingxin Zhou}{bingxin.zhou@sjtu.edu.cnk}

\icmlkeywords{latent diffusion, protein engineering}

\vskip 0.3in
]



\printAffiliationsAndNotice{\icmlEqualContribution} 

\begin{abstract}
The advent of deep learning has introduced efficient approaches for de novo protein sequence design, significantly improving success rates and reducing development costs compared to computational or experimental methods. However, existing methods face challenges in generating proteins with diverse lengths and shapes while maintaining key structural features. To address these challenges, we introduce CPDiffusion-SS, a latent graph diffusion model that generates protein sequences based on coarse-grained secondary structural information. CPDiffusion-SS offers greater flexibility in producing a variety of novel amino acid sequences while preserving overall structural constraints, thus enhancing the reliability and diversity of generated proteins. Experimental analyses demonstrate the significant superiority of the proposed method in producing diverse and novel sequences, with CPDiffusion-SS surpassing popular baseline methods on open benchmarks across various quantitative measurements. Furthermore, we provide a series of case studies to highlight the biological significance of the generation performance by the proposed method. The source code is publicly available at \href{https://github.com/riacd/CPDiffusion-SS}{https://github.com/riacd/CPDiffusion-SS} 
.
\end{abstract}

\section{Introduction}
Deep learning-based protein design provides an innovative and effective methodology, which promotes and creates novel or enhanced functionalities and physical properties of proteins varied from peptides to enzymes. Compared with traditional protein design approaches, such as directed evolution and rational design, deep learning-based protein design can significantly lower the human source, time, and financial cost \cite{chu2024NBT_review2} and create new proteins that do not exist in nature. Protein sequence is the foundation of protein structure and function, indicating that the sequence design is crucial for designing proteins with desired functions. There has been an increasing amount of work on designing protein sequences with deep generative models and validating the effectiveness of the designed protein products through bio-experiments \cite{Chroma2023,zhou2023CPDiffusion}. These new techniques not only offer an opportunity to design novel protein sequences for a protein structure of interest, but also open a new way of designing proteins with significantly enhanced or novel functions for specific biological applications.

\begin{figure*}[t]
    \centering
    \includegraphics[width=\textwidth]{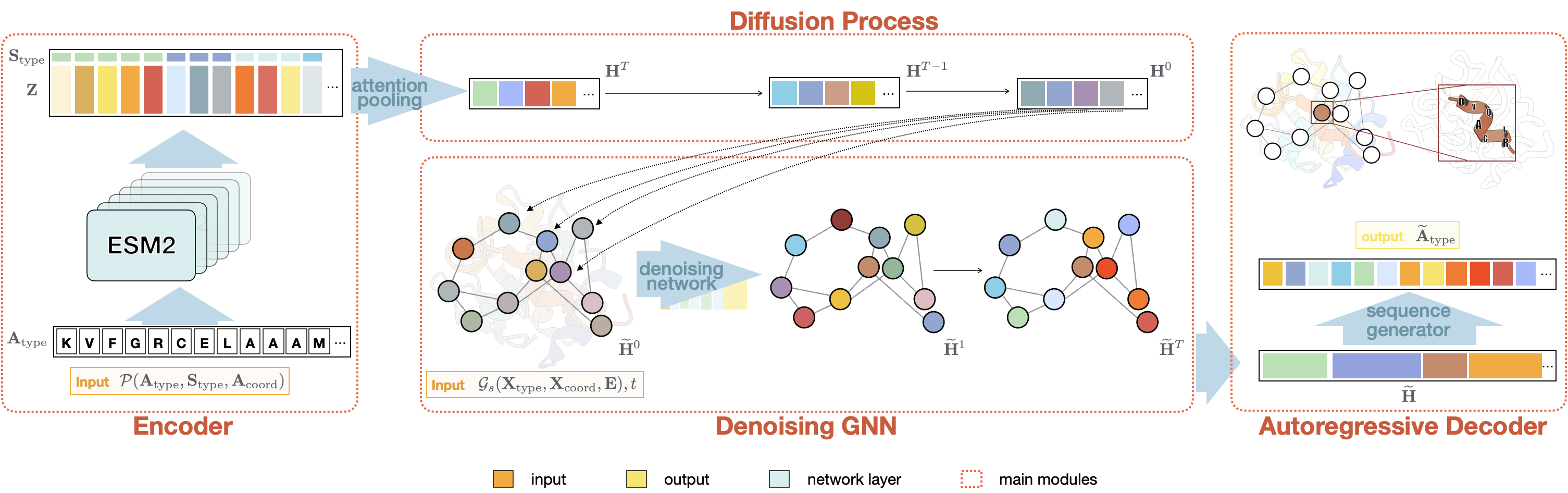}
    \vspace{-3mm}
    \caption{The illustrative figure of \ssdiff. The model embeds AA sequences into a hidden space of secondary structures using the latent graph diffusion model. The generated latent secondary structure representation is then translated into AA sequences of variable lengths by an autoregressive decoder.}
    \label{fig:mainArchitecture}
\end{figure*}

The intricate connection between protein sequences and their functions remains largely unknown due to the vast high-dimensional space of protein sequences. Additionally, obtaining accurately labeled data that detail the sequence-function relationship presents a significant challenge. Thus, the sequence-based deep learning models are generated for finding the relationship between sequence and function. To enhance the generative capabilities, some autoregressive generative models have been developed that incorporate homologous wild-type proteins from closely related functional families or engage multiple sequence alignments. Including protein family data could direct the generated proteins to exhibit specified, desirable traits \cite{poet_neurips2023}. Masked language models adopt a different approach by working with fragments of wild-type protein sequences and training the system to complete the remaining parts \cite{ProtTrans,lin2023ESM2}. Even though protein language models have access to a wealth of sequence data to assimilate typical protein sequence patterns and to craft sequences with variable lengths, it remains a complex task to ensure an ample supply of homologous sequences for specific proteins \cite{rao2021ESM_msa}. A notable shortcoming of these sequence-centric approaches is their tendency to neglect the vital structural features of proteins. These structural elements are critical since they largely dictate protein functionality. Without consideration of these three-dimensional attributes, the models may fail to fully capture the nuances of protein behavior and activity.

Structure-based protein generative models, on the other hand, investigate the conformation of proteins using geometric deep learning models \cite{satorras2021n}. They learn the local interactions of amino acids from the three-dimensional structure of proteins (either from experimental methods \cite{berman2000protein} or folding predictors \cite{lin2023ESM2}) and suggest amino acid compositions for the given scaffold or backbone \cite{dauparas2022ProteinMPNN,yi2024GRADE_IF}. 
Structure-based generative models can learn key patterns of protein composition from fewer data and with smaller model sizes. Moreover, for some structure-driven generative objectives, such as thermostability and binding affinity \cite{zheng2022loosely}, these methods tend to achieve better performance \cite{tan2023protssn}. However, existing structure-based generative methods strictly require knowledge of the exact primary structures of protein for generation. Further, they cannot generate diverse sequences with flexible lengths or based on coarser-grained information, such as the protein's secondary structure.

This study presents \ssdiff, a deep generative model tailored for protein sequence design guided by coarse structural conditions like secondary structure. Such design is valuable for biologists, allowing tailored proteins with specific structural properties. For instance, adjusting $\alpha$-helices or $\beta$-sheets on a protein's surface can enhance its structural rigidity, potentially increasing thermostability \cite{zheng2022loosely}. Moreover, optimizing secondary structures can enhance the encapsulation and delivery efficiency of proteins by viral capsids \cite{yeh2023novo}. Unlike existing methods, \ssdiff~considers protein structure while maintaining flexible amino acid (AA) sequence generation.

We address the challenge of evaluating novel protein sequences generated by deep learning models. Traditionally, quality is assessed based on recovery rate and perplexity compared to wild-type templates \cite{kucera2022ProteoGAN,repecka2021proteinGAN}. However, these metrics have limitations, particularly in evaluating \textit{de novo} design. Instead, we established independent benchmarks based on \textbf{CATH} and proposed new evaluation metrics for assessing novelty, designability, and diversity of novel protein sequences. The empirical evaluation of \ssdiff~includes both quantitative and qualitative analysis. Performance on \textbf{CATH 4.3} dataset surpasses baseline methods in generating secondary structure-based AA sequences. Additionally, case studies suggest its potential applications like enhancing protein functionality and reducing size for drug delivery.

\section{Related Work}
\paragraph{Conditional Protein Sequence Generation}
Protein sequence generation typically seeks to achieve specific catalytic functions, often necessitating the integration of guiding principles or constraints from either the structural or sequence level to obtain the desired results. At the structural level, a common approach is to utilize a fixed protein backbone, such as the positions of amino acids (AAs) in three-dimensional space, and then output the appropriate AA type of each position, forming an AA sequence that is most likely to fold into the given structure \cite{hsu2022ESM_IF}. This approach requires models capable of processing geometric structures, for example, using SE(3) equivariant neural networks to learn the geometric relationships between AAs \cite{satorras2021n}. Open benchmarks have validated these methods for their effectiveness in recovering AAs \cite{dauparas2022ProteinMPNN,yi2024GRADE_IF}. Additionally, in some research, to demonstrate their models' effectiveness in generating desired proteins, wet lab experiments are conducted  \cite{zhou2023CPDiffusion}.
Beyond protein inverse folding, other conditional protein sequence generation tasks require different inputs, such as protein function \cite{kucera2022ProteoGAN}, protein family \cite{repecka2021proteinGAN}, and secondary structure \cite{xie2023helixgan}. Although there have been some methods that attempt to incorporate secondary structures for conditional sequence generation, these methods often have limitations, such as being unable to generate sequences of varying AA length \cite{ni2023ProteinDiffusionGenerator} or secondary structures w fixed order \cite{Chroma2023}.

\paragraph{Protein Language Model}
Protein language models (PLMs) have been a hot spot in the field of AI-assisted protein design. PLMs are trained in a self-supervised manner and utilize extensive amino acid (AA) sequences to extract reliable AA representations, which are valuable for various downstream tasks, such as protein folding \cite{lin2023ESM2} and variant effect prediction \cite{tan2023protssn, poet_neurips2023}. There are two prevalent types of PLMs: masked language models and autoregressive models.
Masked language models are inspired by BERT \cite{devlin2018bert}. These models are trained to predict masked AAs within the context of surrounding unmasked tokens. This approach is exemplified by models like ESM-1b \cite{rives2021ESM-1b}. To improve the model's understanding of sequence characteristics, additional information including evolutionary data from multiple sequence alignments (MSA) \cite{rao2021ESM_msa} or functional annotations \cite{proteinBERT} can be incorporated.
Autoregressive models share an architecture similar to GPT-2 \cite{radford2019gpt2}, generating protein sequences of varying lengths without conditioning \cite{nijkamp2023progen2}. Some PLMs are based on T5 \cite{2020t5}, such as ProtT5 \cite{ProtTrans} and ProstT5 \cite{ProstT5}.


\section{\ssdiff: Secondary Structure-Guided Conditional Latent Protein Diffusion}
In this section, we introduce the problem formulation to our research questions and propose our solution to it, \ie \ssdiff. The notations used in this study is summarized in Table~\ref{tab:notation}.

\begin{table}[t]
    \caption{Table of notations.}
    \label{tab:notation}
    \vspace{2pt}
    \begin{center}
    \resizebox{\linewidth}{!}{
    \begin{tabular}{ll}
    \toprule
    \textbf{notation} & \textbf{description} \\ \midrule
    $\gP(\mA_{\rm type},\mS_{\rm type},\mA_{\rm coord})$ & a protein with sequence and structure information \\
    $\gG_s(\mX_{\rm type},\mX_{\rm coord},\mE)$ & SS-level graph representation of the protein \\
    $\mA_{\rm type}=(\va_1,\dots,\va_n)$ & AA sequence of a protein with $n$ tokens \\
    $\mS_{\rm type}=(\vs_1,\dots,\vs_n)$ & SS label for each AA of a protein \\
    $\mA_{\rm coord}$ & 3D coordinates of each AA in a protein \\
    $\mX_{\rm type}=(\vx_1,\dots,\vx_m)$ & SS sequence of a protein \\
    $\mX_{\rm coord}$ & 3D coordinates of each SS in a protein \\
    $\mZ=[\vz_1,\dots,\vz_n]$ & AA-level latent representation \\
    $\mH = [\vh_1, \vh_2,\dots,\vh_m]$ & SS-level latent representation \\
    $\mH^t$ & latent embeddings at time step $t$ \\
    $\widetilde{\mH}$ & generated SS-level latent representation \\ 
    $\widetilde{\mA}$ & generated AA sequence \\
    $g_{\theta}(\cdot)$ & conditional generative model \\
    $f_{\theta}(\cdot)$ & denoising neural network \\
    \bottomrule\\[-2.5mm]
    \end{tabular}
    }
    \end{center}
\end{table}

\subsection{Problem Formulation}
\label{sec:problemFormulation}
Our study aims to generate AA sequences with secondary structure constraints. Relevant notations can be defined as follows: Let $\gP(\mA_{\rm type},\mS_{\rm type},\mA_{\rm coord})$ denote an arbitrary protein of $n$ AAs, where the two sequences $\mA_{\rm type}=(\va_1,\dots,\va_n)$ and $\mS_{\rm type}=(\vs_1,\dots,\vs_n)$ represent labels for AA types and secondary structure types, respectively. $\mA_{\rm coord}$ is the coordinates of each AA in the three-dimensional Euclidean space. 

The secondary structures (SS) are organized into a graph that illustrates the relationships between them within a protein, denoted as $\gG_s(\mX_{\rm type},\mX_{\rm coord},\mE)$. Here $\mX_{\rm type}$ denotes the sequence of SS type, which can be helix (H), sheet (E), or coil (C).  $\mX_{\rm coord}$ is the coordinates of secondary structures, which is calculated by averaging all AA coordinates within each secondary structure. For instance, in the visualized protein (PDB ID: 1Z25) in Figure~\ref{fig:mainArchitecture}, the highlighted node represents a helix structure, containing $7$ AAs in the wild-type template. Suppose the structure is located at $j$-th position in  $\mX_{\rm type}$ and the $7$ corresponding AAs are located sequentially starting from the $i$-th position in $\mA_{\rm type}$, then the coordinates are computed as $\mX_{\rm coord;j} = {\rm mean}(\mA_{\rm coord;i},\dots,\mA_{\rm coord;i+6})$. Then, the SSs are connected to their $k$ nearest neighbor in the Euclidean space, with edge features $\mE$ encoding the Euclidean distance between each SS pair. 

The objective is to train a conditional generative model $g_{\theta}(\cdot)$ which generates the desired AA sequence $\widetilde{\mA}=(\tilde{\va}_1,\dots,\tilde{\va}_{n^{\prime}})$, where $n^{\prime}$ does not necessarily equals $n$, \ie
\begin{equation}
    \widetilde{\mA} = g_{\theta}\left(\gG_s(\mX_{\rm type},\mX_{\rm coord},\mE)\right).
\end{equation}
The major challenge is that only coarse information about the desired structures is provided. Conventional protein language models and inverse folding methods are inadequate: protein language models cannot directly constrain the structure, and inverse folding methods require precise structural inputs, including the exact number and positions of all AAs. To address this, we propose \ssdiff, a secondary structure-guided conditional latent protein diffusion method for approximating $g_{\theta}(\cdot)$.

\subsection{Model Architecture}
\ssdiff~comprises three components: a sequence encoder, a latent diffusion generator, and an autoregressive decoder. The encoder and decoder form a variational auto-encoder. The sequence encoder embeds amino acid (AA) sequences into a latent space characterized by secondary structure-level (SS-level) representations, while the decoder translates these SS-level latent representations back to the AA space. Both the encoder and decoder use protein language models for sequence embedding and reconstruction. The central component, a latent graph diffusion model, generates diverse SS-level hidden representations within the latent space conditioned on SS input. Below, we detail the construction of each module.

\subsubsection{Encoder-Decoder}
\paragraph{Encoder}
For a protein $\gP$ including $n$ AAs and $m$ SSs, the encoder converts the discrete input AA sequence $(\va_1, \ldots, \va_n)$ into a continuous representation sequence $(\vh_1, \ldots, \vh_m)$ using a protein language model and an attention pooling module. The pre-trained protein language model initially maps the AA sequences of proteins to AA-level vector representations $\mZ=[\vz_1, \ldots, \vz_n]$. In this process, we utilize an evolutionary-scale protein language model \cite{lin2023ESM2} to effectively analyze the structural and functional characteristics of proteins, employing a masked language model training objective \cite{devlin2018bert}, \ie 
\begin{equation*}
    \Ls_{\rm {MLM}}:= -\sum_{i \in \mM}\log\left(\sP(\rva_i|\rmA_{\backslash\mM})\right),
\end{equation*}
where  $\rmA_{\backslash\mM}$ represents the masked AA sequence obtained from $\mA_{\rm type}$. 
To obtain secondary structure (SS)-level representations, we utilize an attention pooling module \cite{yang2023MIFST}, which aggregates amino acid (AA)-level representations $\mZ$ into SS-level representations $\mH = [\vh_1, \vh_2, \dots, \vh_m]$. Using $\mX_{\rm type}$, we rearrange $\mZ$ into $m$ groups:
\begin{equation*}
    \mZ = [\mZ_1,\dots,\mZ_m] = \left[[\vz_1,\dots,\vz_{n_1}],\dots, [\vz_{n-n_m+1},\dots,\vz_{n}]\right],
\end{equation*} 
with $n_i$ being the number of AA in the $i$-th secondary structure, \ie $\sum_{i=1}^{m}{n_i}=n$. For the $k$-th ($1\leq k\leq m$) secondary structure, the corresponding latent embedding $\vh_k$ is summarized from 
$\mZ_k$ by
\begin{equation}
    \vh_k={\rm AttnPool}(\mZ_k) = \softmax(\mathrm{Conv}(\mZ_k))\cdot\mZ_k,
\end{equation}
where $\mathrm{Conv}(\cdot)$ represents a 1-dimensional convolution along the dimension of the AA sequence and $\cdot$ calculates the weighted average of AA embeddings within the same secondary structure.

\paragraph{Decoder}
The decoder converts the diffusion-generated SS-level representation (introduced in the following section) $\widetilde{\mH}=(\tilde{\vh}_1,\dots,\tilde{\vh}_{m})$ into $\widetilde{\mA}=(\tilde{\va}_1,\dots,\tilde{\va}_{n^{\prime}})$. To generate AA sequences of varying lengths, an autoregressive model with multi-layer cross-attention is employed \cite{vaswani2017attention}. The learning objective is structured as a sequence translation task. For training, the SS-level hidden representation $(\vh_1, \dots, \vh_m)$ from the encoder is used. This continuous representation is fed into the decoder as context vectors, guiding the reconstruction of the AA sequence. The decoder's training target is to minimize the KL divergence.
\begin{equation}
\label{eq:decoder_kl}
\min\sum_{\va_i\in\mA}\KL\left(\va_i||{\rm Decoder}({\rm Encoder}(\mA), \va_{<i})\right).
\end{equation}
Rotary Position Embedding (RoPE) \cite{su2024RoPE} is applied for positional encoding of the AA sequences, enhancing the model's ability to effectively capture positional information.

In summary, the encoder-decoder mechanism facilitates the mapping between AA-level protein sequences and SS-level latent space. We utilize an evolutionary model to proficiently perform sequence embedding and train an autoregressive decoder for translating AA sequences of varying lengths. To better align with secondary structure conditions and enrich the diversity of generated outcomes, we incorporate latent graph diffusion to generate SS-level vector representations. 

\begin{figure}[t]
    \centering
    \includegraphics[width=\linewidth]{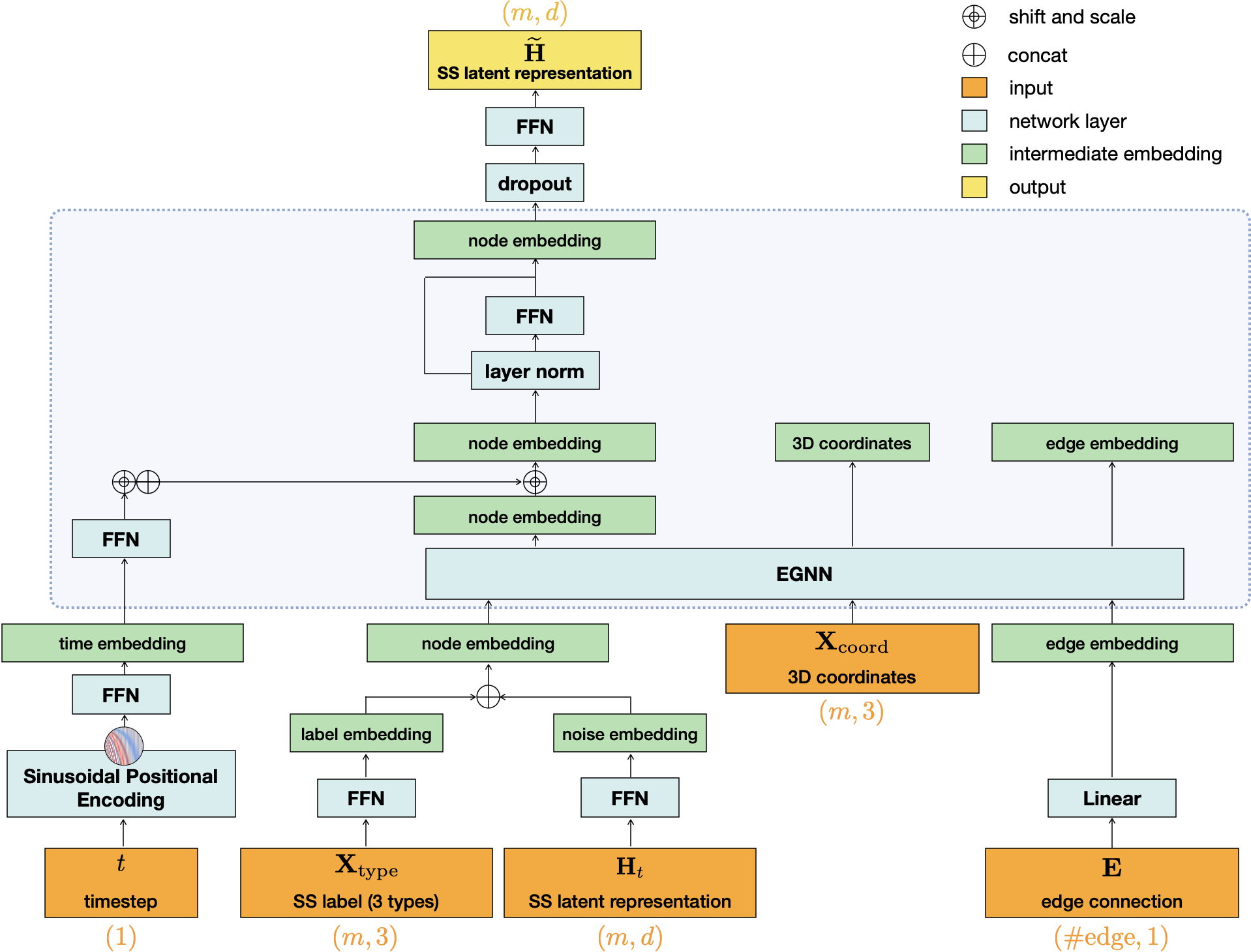}
    \vspace{-2mm}
    \caption{Illustrative architecture of the latent diffusion model.}
    \label{fig:architecture_diffusion}
\end{figure}

\subsubsection{Latent Diffusion}
For generating SS-level latent representations, we adhere to the standard pipeline of the denoising diffusion probabilistic model \cite{ho2020DDPM} within the latent space. For each protein AA sequence, we extract its secondary embeddings from the pre-trained encoder, denoted as $\mH = [\vh_1, \vh_2, ..., \vh_m]$. The diffusion model is trained to generate representations of protein sequences that adhere to the secondary structure properties. The architecture for the denoising model is visualized in Figure~\ref{fig:architecture_diffusion}.

Following the denoising diffusion probabilistic model (DDPM), the forward diffusion process gradually adds Gaussian noise to the input embeddings over steps $0 \rightarrow T$. 
The objective is to maximize the evidence lower bound, which is equivalent to minimizing the expected reconstruction loss
\begin{align*}
    \min_{\theta} \mathbb E_{t, \mH^t} \| f_\theta(\mH^t, t, \mX_{\rm coord}, \mX_{\rm type}, \gG_s) - \mH^0 \|
\end{align*}
where we ignore the weighting constants. 

To incorporate conditions based on secondary structure information, we design the denoising neural network $f_\theta(\cdot)$ using equivariant graph neural networks \cite{satorras2021n}. Each protein is represented as an SS-level graph $\gG_s(\mX_{\rm type}, \mX_{\rm coord}, \mE)$, preserving the 3D geometric information of the secondary structures in $\mX_{\rm coord}=[\vx_1, ...,\vx_m]$. As previously introduced, these coordinates are defined by the average $C_{\alpha}$ of the corresponding AAs within each secondary structure. In addition to the positions of the secondary structures, their types are encoded using one-hot encoding features $\mX_{\rm type}$.

The denoising network $f_\theta(\cdot)$ is conditioned on the 3D positions of protein secondary structures, thus the predicted embeddings should be invariant to orthogonal transformations or translations of the input coordinates. This means that translating, reflecting, or rotating the input should result in equivalent transformations of the output. To achieve this, we use $E(3)$ equivariant graph neural networks (EGNN) \cite{satorras2021n} as the backbone for $f_\theta(\cdot)$. EGNNs have proven effective for protein representation learning \cite{tan2023protssn,yi2024GRADE_IF,zhou2023CPDiffusion}. At the $\ell$-th layer, the hidden representation $\vh_i^{\ell+1}$ is updated by edge and position updates, followed by node aggregation: 
\begin{align*}
    \vm_{ij}^{\ell+1} &= \phi_e(\vh_i^\ell, \vh_j^\ell, \| \vx_i - \vx_j \|^2, \ve_{ij}) \\
    \vx_i^{\ell+1} &= \vx_i^\ell + \sum_{j \neq i} (\vx_i^\ell - \vx_j^\ell) \phi_x(\vx_i^\ell) \\
    \vh_i^{\ell + 1} &= \phi_h( \vh_i^\ell , \sum_{j\neq i}, \vm_{ij}),
\end{align*}
where $\phi_e(\cdot)$ and $\phi_h(\cdot)$ are the edge and node propagation functions, respectively, and $\ve_{ij}$ represents the edge feature between nodes $i$ and $j$.

\begin{table*}[!t]
\caption{Diversity, novelty, and consistency of secondary-structure-guided generation by baseline methods on $50$ test protein templates structures from \textbf{CATH4.3}. For each measurement, we report the average score with the standard deviation in parentheses. The best performance for each metric is indicated in \bm{$bold$}, while the second-best performance is \underline{$underlined$}.}
\label{tab:result_main}
\vspace{1mm}
\begin{center}
\resizebox{\linewidth}{!}{
    \begin{tabular}{lcccccccccccc}
    \toprule
    &\multicolumn{3}{c}{\textbf{Diversity}} & \textbf{Novelty} & \multicolumn{3}{c}{\textbf{Consistency} (SS3)} & \multicolumn{3}{c}{\textbf{Consistency} (SS3, w/o loop)} \\ \cmidrule(lr){2-4}\cmidrule(lr){5-5}\cmidrule(lr){6-8}\cmidrule(lr){9-11}
    & TM~{\tiny new} $\downarrow$ & RMSD $\uparrow$ & Seq. ID $\downarrow$ & TM~{\tiny wt}$\downarrow$ & ID  $\uparrow$ & ID~{\tiny max} $\uparrow$ & MSE~{\tiny SS Composition} $\downarrow$ & ID  $\uparrow$ & ID~{\tiny max} $\uparrow$ & MSE~{\tiny SS Composition} $\downarrow$ \\
    \midrule
    \textsc{vanilla decoder} & \underline{$0.27 \pm 0.01$} & $3.98 \pm 0.22$ & \bm{$6.31 \pm 0.12$} & $0.23 \pm 0.11$ & $66.28 \pm 11.47$ & $76.80 \pm 11.64$ & $6.16 \pm 3.08$ & $56.82 \pm 15.46$ & $70.31 \pm 15.30$ & $23.59 \pm 13.95$ \\[1mm]
    \textsc{ProstT5} & $0.28 \pm 0.03$ & \bm{$6.65 \pm 1.01$} & $15.78 \pm 2.31$ & \bm{$0.12 \pm 0.06$} & {$74.31 \pm 9.98$} & {$82.52 \pm 12.35$} & $4.19 \pm 2.00$ & $66.61 \pm 14.00$ & $77.24 \pm 15.98$ & $17.32 \pm 9.11$ \\[1mm]
    \textsc{ESM2} (1) & \bm{$0.26 \pm 0.06$} & $3.15 \pm 1.11$ & $19.48 \pm 6.00$ & $0.29 \pm 0.15$ & $45.25 \pm 14.67$ & $53.98 \pm 17.02$ & $7.31 \pm 3.24$ & $35.17 \pm 19.75$ & $44.13 \pm 21.8$ & $29.26 \pm 16.36$\\[1mm]
    \textsc{ESM2} (0.8) & {$0.27 \pm 0.04$} & $3.44 \pm 0.96$ & $13.02 \pm 2.52$ & $0.30 \pm 0.15$ & $52.41 \pm 20.58$ & $58.23 \pm 20.75$ & $6.44 \pm 3.91$ & $41.50 \pm 27.19$ & $50.08 \pm 25.99$ & $28.32 \pm 18.66$ \\ [1mm]
    \textsc{ESM-if1}  & $0.29 \pm 0.01$ & {$4.97 \pm 0.85$} & {$7.46 \pm 0.61$} & $0.20 \pm 0.09$ & \underline{$78.53 \pm 10.27$} & \underline{$84.88 \pm 10.13$} & \underline{$3.34 \pm 1.93$} & \underline{$74.44 \pm 11.83$} & \underline{$82.55 \pm 11.92$} & \underline{$14.91 \pm 9.01$} \\ [1mm]
    \textsc{ProteinMPNN}  & $0.35 \pm 0.19$ & $4.47 \pm 1.58$ & $76.50 \pm 17.12$ & $0.19 \pm 0.14$ & $56.00 \pm 22.09$ & $62.56 \pm 21.66$ & $6.46 \pm 4.16$ & $47.28 \pm 26.33$ & $53.73 \pm 27.38$ & {$17.77 \pm 13.32$} \\ [1mm]
    \midrule
    \ssdiff & $0.30 \pm 0.02$ & \underline{$5.69 \pm 0.78$} & \underline{$7.08 \pm 0.36$} & \underline{$0.16 \pm 0.07$} & \bm{$81.57 \pm 9.78$} & \bm{$86.95 \pm 9.75$} & \bm{$1.56 \pm 0.89$} & \bm{$77.84 \pm 12.93$} & \bm{$84.43 \pm 12.05$} & \bm{$6.61 \pm 3.86$} \\
    \bottomrule\\[-1.5mm]
    \end{tabular}
}
\end{center}
\end{table*}

\subsection{Model Pipeline}
\paragraph{Data Preparation}
\ssdiff\ utilizes two types of protein data as model input: the AA-level protein representation $\gP(\mA_{\rm type}, \mS_{\rm type}, \mA_{\rm coord})$ and the SS-level graph representation $\gG_s(\mX_{\rm type}, \mX_{\rm coord}, \mE; \mH)$, as previously discussed in Section~\ref{sec:problemFormulation}. For $\gP$, both $\mA_{\rm type}$ and $\mA_{\rm coord}$ are directly obtained from structure-informed protein documents, such as PDB. The secondary structure $\mS_{\rm type}$ is assigned using DSSP \cite{touw2015dssp_2}. Proteins with more than 100 AAs in a single secondary structure are excluded, as they are believed to be problematic or irregularly dominated by loops. The processed AA-level data $\gP$ is used solely for training purposes.
In contrast, SS-level information $\gG_s$ is used for both training and inference. Constructing the associated graph representation requires additional data processing steps. Specifically, the SS-level graph for a protein is defined with each node representing a secondary structure, labeled using one-hot encoding for its class (H, E, or C). Additionally, each secondary structure has a 1280-dimensional hidden representation $\mH$ from the encoder that describes its AA compositions. During inference, this feature is generated by latent diffusion and represented as $\widetilde{\mH}$. The three-dimensional coordinate $\mX_{\rm coord}$ is defined as the average position of all AAs (determined by the $C_{\alpha}$ atom) within it.
Following the convention for constructing protein graphs, the connections in $\gG_s$ are defined using $k$-nearest neighbor ($k$NN) graphs, with $k=3$ based on the fact that secondary structures are less closely related than AAs. The edge matrix $\mE$ is weighted by the fraction of the Euclidean distance between connected node pairs.

\paragraph{Model Training and Inferencing}
\ssdiff~undergoes a two-stage training process, with separate training phases for the encoder-decoder module and the latent diffusion module. For the encoder-decoder, we utilize the pre-trained ESM2-650M \cite{lin2023ESM2} and train our Transformer-style decoder to minimize the objective function described in (\ref{eq:decoder_kl}). This model is trained on a subset of the \textbf{AlphaFoldDB} \cite{barrio2023AFDBclustering}\footnote{Available at \url{https://alphafold.ebi.ac.uk/}} clustered by FoldSeek \cite{van2024Foldseek}, which includes over 2 million wild-type proteins with \textsc{AlphaFold2} predictions.
In the second stage, we freeze the trained encoder-decoder and train the latent graph diffusion model to reconstruct the latent secondary structure representation $\mH$. Given that the performance of the latent graph diffusion is closely related to protein structure, we train the model using \textbf{CATH4.3} \cite{sillitoe2021cath}, which provides over 30,000 protein domain structures with less than 40
The inference process begins with the latent diffusion model, which uses the provided secondary structure graph and a randomly generated noise representation $\mH^T$. It then proceeds through the denoising process using EGNN layers to generate latent representations conditioned on the specified input secondary structure. Subsequently, the sampled $\widetilde{\mH}$ is fed into the trained decoder to translate each secondary structure representation into explicit AA sequences.

\section{Experiments}

\subsection{Experimental Protocol}

\paragraph{Generation Task}
The models are evaluated through a secondary structure-based protein sequence generation task. We use $50$ randomly selected structure templates from \textbf{CATH4.3} for validation. These $50$ test templates are excluded from the training set to ensure unbiased evaluation. For each template, $200$ sequences are generated and assessed based on their structures predicted by \textsc{ESMFold2}. Models are provided with the secondary structure and the minimum essential additional data required for each specific baseline model.

Specifically, for any given template structure, \ssdiff~receives an SS-level graph featuring secondary structure labels. Structure-based models (\textsc{ProteinMPNN} and \textsc{ESM-if1}) receive AA-level graph representations, where amino acids (AAs) within the same secondary structure are positioned at the group's center. Alternatively, sequence-based methods (\textsc{ESM2} \cite{lin2023ESM2} and \textsc{ProstT5} \cite{2020t5}) are given a small set of unmasked AA tokens. Both \textsc{ProstT5} and ESM2 (1) obtain a randomly selected unmasked AA token in each secondary structure, while ESM2 (0.8) and ESM2 (0.6) are supplied with randomly selected $20\%$ and $40\%$ unmasked AAs from the wild-type protein, respectively. We exclude a comparison with the model described in \cite{ni2023ProteinDiffusionGenerator} due to the unavailability of the model implementation's checkpoint.

\paragraph{Training Setup}
For the encoder module, we utilize ESM-650, followed by a convolutional 1D-attention mechanism. The input channel for the convolution operator is set to $1280$ (the output dimension of ESM2-650M), with the output channel being $1$ and a kernel size of $1$.
In the latent graph diffusion module, we employ $4$ EGNN layers as the denoising layers. The hidden and embedding dimensions are set to $640$ and $1280$, respectively. We use the \texttt{sqrt} noise schedule, with a learning rate of $5 \times 10^{-4}$ and a weight decay of $10^{-5}$.
For the decoder, we incorporate $3$ Transformer layers, each with $8$ attention heads and hidden dimensions of $4960$ in the feed-forward network.
All implementations are programmed using \texttt{PyTorch Geometric} (version 2.4.0) \citep{fey2019fast} and \texttt{PyTorch} (version 2.2). The training is conducted on 8 NVIDIA\textsuperscript{®} Tesla A800 GPUs, each with $80$GB HBM2, mounted on an HPC cluster.
To ensure reproducibility, all details required to replicate our results are included in the submission.

\subsection{Evaluation Measurements}

\paragraph{Diversity} assesses the variance of generated amino acid (AA) sequences from the same template structure. We evaluate the diversity of the generated results at both the sequence and structure levels by comparing the pairwise similarity of all generated sequences and reporting the average scores.
For sequence-level evaluation, we calculate the AA sequence identity, expressed as a percentage. For structure-level evaluation, we use TM-score and RMSD (Root Mean Square Deviation), both calculated using TM-align \cite{zhang2005tm-align}. These metrics are crucial as we aim for generated sequences from the same template to exhibit significant differences. Thus, we prefer models that generate sequences with lower average sequence identity, lower average TM-score, and higher average RMSD. In Table~\ref{tab:result_main}, these measurements are denoted as \textit{Seq. ID}, \textit{TM~{\tiny new}}, and \textit{RMSD}, respectively.

\paragraph{Novelty} evaluates whether the structures of generated proteins significantly differ from existing wild-type proteins. Maximizing novelty is a common design objective in \textit{de novo} protein design \cite{watson2023RFDiffusion, yim2024SE3flow_match}.
For efficient protein structure comparison, we use Foldseek \cite{van2024Foldseek} to examine the alignment of the generated protein structures (predicted by ESMFold) with those in the training set. We report the TM-score between the most similar wild-type protein and the generated protein. In this context, a lower TM-score indicates higher novelty, which is desirable.
The novelty evaluation is reported under \textit{TM~{\tiny wt}}. We provide the average TM-scores for all the proteins generated from the test templates.

\paragraph{Consistency} evaluates the alignment between the input secondary structure conditions and the predicted secondary structure of generated proteins. This is measured from two perspectives: SS-level sequence identity and structure composition.
Similar to AA-level sequence identity, SS-level sequence identity is computed by aligning sequences and calculating the proportion of matched tokens. The best-aligned sequence is obtained by maximizing the alignment length corresponding to the secondary structure sequence alignment, using a penalty mechanism for mismatches and gaps. This follows the definition of AA sequence identity in BLAST \cite{altschul1990blast} for global sequence comparison, where ${\rm identity} = ({\rm Matches} / {\rm Alignment Length})\times 100\%.$
The alignment length includes the total number of tokens for matches, gaps, and mismatches.
Secondary structure composition calculates the proportions of helices (H), sheets (E), and coils (C) in both the input condition and the generated sequences. It then employs the Mean Squared Error (MSE) measure to quantify their differences.
The three introduced metrics are reported in Table~\ref{tab:result_main} as \textit{ID}, \textit{ID~{\tiny max}}, and \textit{MSE~{\tiny SS Composition}}. Additionally, since sheets and helices are generally considered more important and harder to generate than loops (coils), and their structures are more fixed, we also report the consistency score after removing loops as a reference.

\begin{figure}[t]
    \centering
    \includegraphics[width=\linewidth]{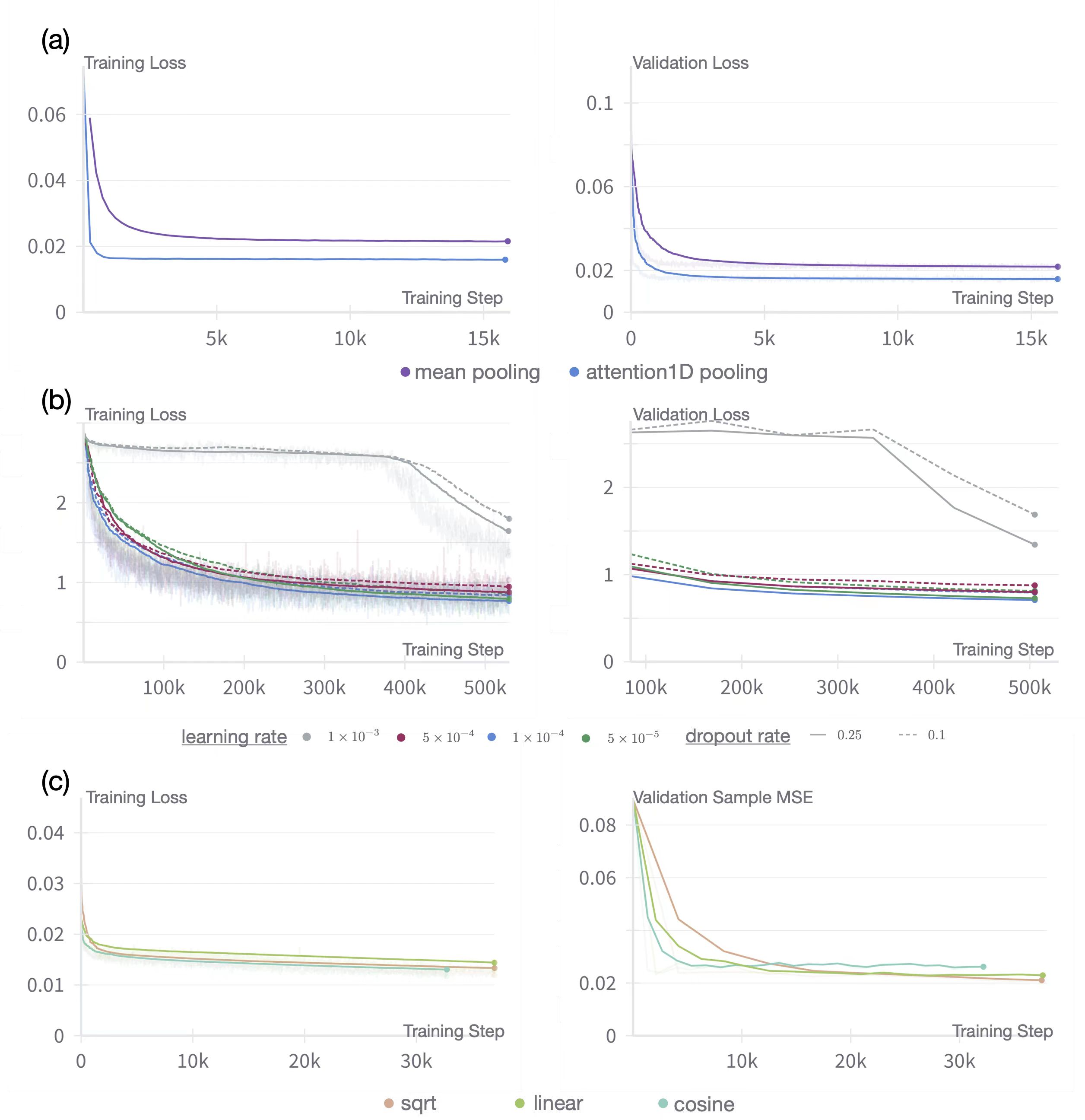}
    \vspace{-5mm}
    \caption{Learning curve with different (a) pooling layers; (b) learning rate and dropout rate. (c) noise schedules in the diffusion model.}
    \label{fig:learningCurve}
\end{figure}

\begin{figure*}[t]
    \centering
    \includegraphics[width=\textwidth]{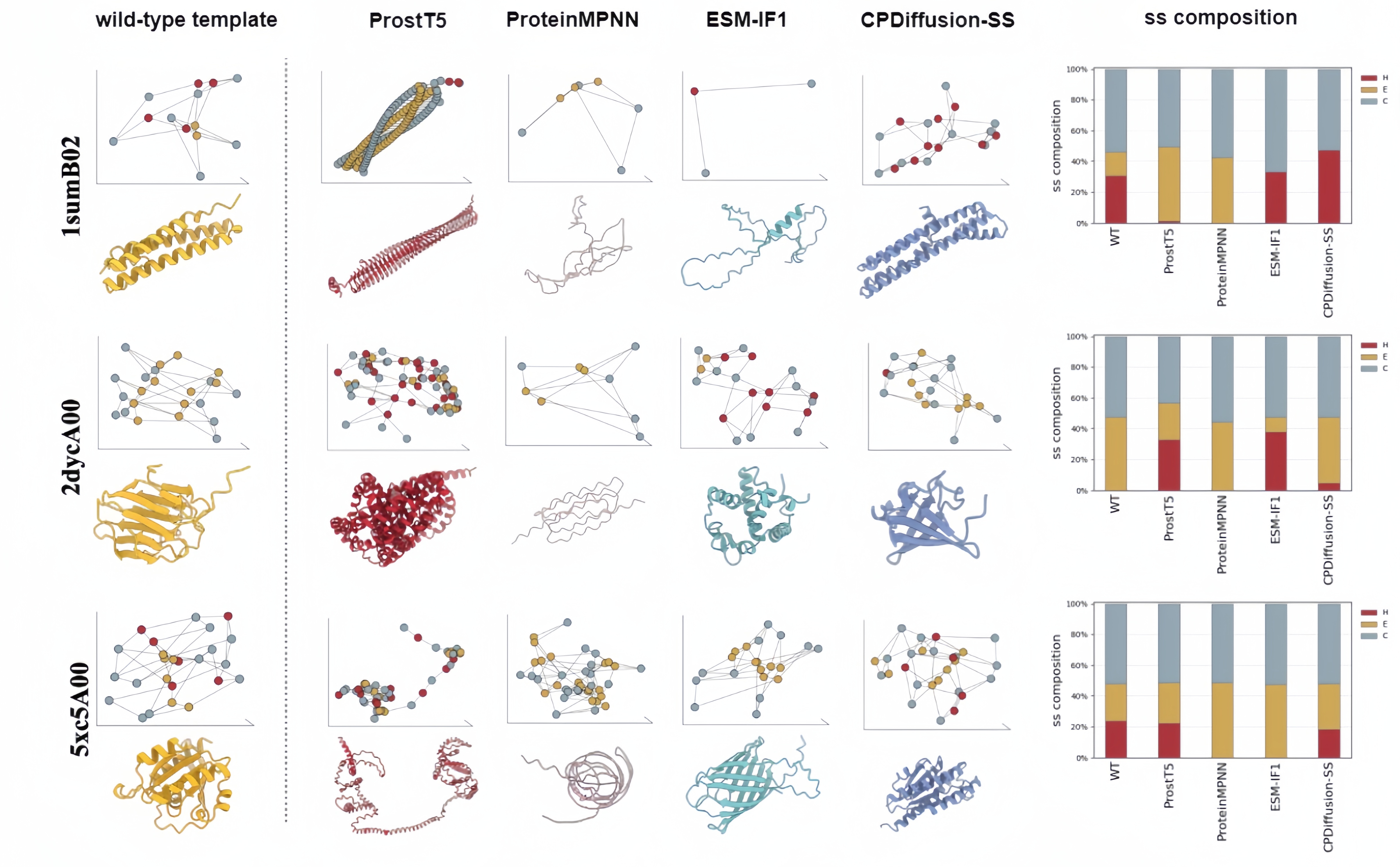}
    \vspace{-3mm}
    \caption{Predicted 3D structures and composition of secondary structures on three cases from the test dataset. Here we use red, yellow, and blue colors to represent helices (H), sheets (E), and coils (C), respectively.
    }
    \label{fig:caseStudy}
\end{figure*}

\subsection{Results Analysis}
The generative performance scores are presented in Table~\ref{tab:result_main}, where we assess our proposed \ssdiff~against both sequence and structure-based baseline methods across $10$ evaluation metrics focusing on the diversity, novelty, and consistency of the generated sequences. In this evaluation, \ssdiff~outperforms baseline methods on $9$ out of the 10 metrics, except for TM~{\tiny new}, where language models generally exhibit superior performance compared to structure-aware models. This disparity can be attributed to language models not explicitly integrating structural information, thus allowing for more unrestricted sequence generation. For instance, sequences generated by \textsc{ProstT5} for all three examined templates frequently exhibit repetitive patterns of certain amino acids, such as Glycine, Leucine, and Isoleucine. However, these amino acids are commonly found in all proteins for backbone stabilization and lack specificity to individual proteins. Moreover, such sequences are highly improbable to occur naturally, leading to lower sequence identity scores and TM scores.

Additionally, language models tend to produce longer sequences compared to structure-constrained models like ESM-if1 and \ssdiff. This observation is evident in Figure~\ref{fig:caseStudy}, where ProstT5 generates significantly longer sequences compared to both baseline methods and the wild-type templates.

Furthermore, we analyze the learning curve of the trained model and compare it with other hyperparameter configurations, as illustrated in Figure~\ref{fig:learningCurve}. All curves are visualized using \texttt{wandb} with moving average smoothing for better clarity. Both training and validation curves rapidly converge to a stable state after a reasonable number of training steps. To justify our choice of hyperparameters and architectures, we compare the learning curve with different pooling methods (average pooling and attention pooling), learning rates, and dropout rates for the encoder-decoder, as well as noise schedules (sqrt, linear, and cosine) for the latent diffusion model.

\subsection{Case Study}
To demonstrate \ssdiff's efficacy in using secondary structures for protein sequence generation, we conducted experiments generating novel sequences guided by specific secondary structures. We selected protein structures shown in Figure~\ref{fig:caseStudy}(a) as constraints. We then compared the structures of sequences generated by \ssdiff, \textsc{ProstT5}, \textsc{ESM-if1}, and \textsc{ProteinMPNN} under the same conditions. Sequences from \ssdiff~fold into plausible protein structures with similar secondary structure compositions to the wild-type template. In contrast, sequences from \textsc{ProstT5}, \textsc{ESM-if1}, and \textsc{ProteinMPNN} deviate significantly from the secondary structural conditions. Notably, \textsc{ProstT5} generates sequences much longer than the template, and \textsc{ProteinMPNN} produces sequences forming only random coils, unlikely to fold into functional proteins.



\section{Conclusion and Discussion}
This study introduces a novel protein generation model guided by secondary structures, crucial elements for protein functionality. Leveraging powerful protein language models and latent graph diffusion models, we develop one of the first deep learning frameworks capable of generating diverse and reliable sequences conditioned on specific secondary structures.

Our experimental findings underscore \ssdiff's ability to generate proteins with target structures while adhering to secondary structure constraints. This capability holds significant implications for protein design and protein-based biotechnology. Structural flexibility, crucial for protein stability and activity, is intricately linked to secondary structure. Proteins often encounter challenges in industrial applications within extreme environments like strong acids, bases, or high temperatures due to structural instability. \ssdiff~offers a solution by introducing new helices and sheets on the protein surface, compacting the protein and enhancing its resistance to extreme conditions \cite{zheng2022loosely}. Additionally, the flexibility of a protein's catalytic pocket profoundly influences its bioactivity \cite{zheng2022loosely}. By employing \ssdiff~to increase loops and turns around catalytic sites, conformational changes can be facilitated during biofunctions, thereby enhancing catalytic activity.

\section*{Impact Statement}
This paper presents work whose goal is to advance the field of 
protein de novo design. There are many potential societal consequences 
of our work, none of which we feel must be specifically highlighted here.

\bibliography{reference}

\begin{thebibliography}{39}
\providecommand{\natexlab}[1]{#1}
\providecommand{\url}[1]{\texttt{#1}}
\expandafter\ifx\csname urlstyle\endcsname\relax
  \providecommand{\doi}[1]{doi: #1}\else
  \providecommand{\doi}{doi: \begingroup \urlstyle{rm}\Url}\fi

\bibitem[Altschul et~al.(1990)Altschul, Gish, Miller, Myers, and Lipman]{altschul1990blast}
Altschul, S.~F., Gish, W., Miller, W., Myers, E.~W., and Lipman, D.~J.
\newblock Basic local alignment search tool.
\newblock \emph{Journal of molecular biology}, 215\penalty0 (3):\penalty0 403--410, 1990.

\bibitem[Barrio-Hernandez et~al.(2023)Barrio-Hernandez, Yeo, J{\"a}nes, Mirdita, Gilchrist, Wein, Varadi, Velankar, Beltrao, and Steinegger]{barrio2023AFDBclustering}
Barrio-Hernandez, I., Yeo, J., J{\"a}nes, J., Mirdita, M., Gilchrist, C.~L., Wein, T., Varadi, M., Velankar, S., Beltrao, P., and Steinegger, M.
\newblock Clustering predicted structures at the scale of the known protein universe.
\newblock \emph{Nature}, 622\penalty0 (7983):\penalty0 637--645, 2023.

\bibitem[Berman et~al.(2000)Berman, Westbrook, Feng, Gilliland, Bhat, Weissig, Shindyalov, and Bourne]{berman2000protein}
Berman, H.~M., Westbrook, J., Feng, Z., Gilliland, G., Bhat, T.~N., Weissig, H., Shindyalov, I.~N., and Bourne, P.~E.
\newblock The protein data bank.
\newblock \emph{NAR}, 28\penalty0 (1):\penalty0 235--242, 2000.

\bibitem[Brandes et~al.(2022)Brandes, Ofer, Peleg, Rappoport, and Linial]{proteinBERT}
Brandes, N., Ofer, D., Peleg, Y., Rappoport, N., and Linial, M.
\newblock {ProteinBERT: a universal deep-learning model of protein sequence and function}.
\newblock \emph{Bioinformatics}, 38\penalty0 (8):\penalty0 2102--2110, 02 2022.
\newblock ISSN 1367-4803.

\bibitem[Chu et~al.(2024)Chu, Lu, and Huang]{chu2024NBT_review2}
Chu, A.~E., Lu, T., and Huang, P.-S.
\newblock Sparks of function by de novo protein design.
\newblock \emph{Nature Biotechnology}, 42\penalty0 (2):\penalty0 203--215, 2024.

\bibitem[Dauparas et~al.(2022)Dauparas, Anishchenko, Bennett, Bai, Ragotte, Milles, Wicky, Courbet, de~Haas, Bethel, et~al.]{dauparas2022ProteinMPNN}
Dauparas, J., Anishchenko, I., Bennett, N., Bai, H., Ragotte, R.~J., Milles, L.~F., Wicky, B.~I., Courbet, A., de~Haas, R.~J., Bethel, N., et~al.
\newblock Robust deep learning--based protein sequence design using proteinmpnn.
\newblock \emph{Science}, 378\penalty0 (6615):\penalty0 49--56, 2022.

\bibitem[Devlin et~al.(2018)Devlin, Chang, Lee, and Toutanova]{devlin2018bert}
Devlin, J., Chang, M.-W., Lee, K., and Toutanova, K.
\newblock Bert: Pre-training of deep bidirectional transformers for language understanding.
\newblock \emph{arXiv:1810.04805}, 2018.

\bibitem[Elnaggar et~al.(2021)Elnaggar, Heinzinger, Dallago, Rehawi, Yu, Jones, Gibbs, Feher, Angerer, Steinegger, Bhowmik, and Rost]{ProtTrans}
Elnaggar, A., Heinzinger, M., Dallago, C., Rehawi, G., Yu, W., Jones, L., Gibbs, T., Feher, T., Angerer, C., Steinegger, M., Bhowmik, D., and Rost, B.
\newblock Prottrans: Towards cracking the language of lifes code through self-supervised deep learning and high performance computing.
\newblock \emph{IEEE TPAMI}, pp.\  1--1, 2021.
\newblock \doi{10.1109/TPAMI.2021.3095381}.

\bibitem[Fey \& Lenssen(2019)Fey and Lenssen]{fey2019fast}
Fey, M. and Lenssen, J.~E.
\newblock Fast graph representation learning with {PyTorch Geometric}.
\newblock In \emph{ICLR Workshop on RLGM}, 2019.

\bibitem[Heinzinger et~al.(2023)Heinzinger, Weissenow, Sanchez, Henkel, Steinegger, and Rost]{ProstT5}
Heinzinger, M., Weissenow, K., Sanchez, J.~G., Henkel, A., Steinegger, M., and Rost, B.
\newblock Prostt5: Bilingual language model for protein sequence and structure.
\newblock \emph{bioRxiv}, 2023.
\newblock \doi{10.1101/2023.07.23.550085}.

\bibitem[Ho et~al.(2020)Ho, Jain, and Abbeel]{ho2020DDPM}
Ho, J., Jain, A., and Abbeel, P.
\newblock Denoising diffusion probabilistic models.
\newblock \emph{NeurIPS}, 33:\penalty0 6840--6851, 2020.

\bibitem[Hsu et~al.(2022)Hsu, Verkuil, Liu, Lin, Hie, Sercu, Lerer, and Rives]{hsu2022ESM_IF}
Hsu, C., Verkuil, R., Liu, J., Lin, Z., Hie, B., Sercu, T., Lerer, A., and Rives, A.
\newblock Learning inverse folding from millions of predicted structures.
\newblock In \emph{ICML}, pp.\  8946--8970. PMLR, 2022.

\bibitem[Ingraham et~al.(2023)Ingraham, Baranov, Costello, Barber, Wang, Ismail, Frappier, Lord, Ng-Thow-Hing, Van~Vlack, Tie, Xue, Cowles, Leung, Rodrigues, Morales-Perez, Ayoub, Green, Puentes, Oplinger, Panwar, Obermeyer, Root, Beam, Poelwijk, and Grigoryan]{Chroma2023}
Ingraham, J.~B., Baranov, M., Costello, Z., Barber, K.~W., Wang, W., Ismail, A., Frappier, V., Lord, D.~M., Ng-Thow-Hing, C., Van~Vlack, E.~R., Tie, S., Xue, V., Cowles, S.~C., Leung, A., Rodrigues, J.~a.~V., Morales-Perez, C.~L., Ayoub, A.~M., Green, R., Puentes, K., Oplinger, F., Panwar, N.~V., Obermeyer, F., Root, A.~R., Beam, A.~L., Poelwijk, F.~J., and Grigoryan, G.
\newblock Illuminating protein space with a programmable generative model.
\newblock \emph{Nature}, 2023.
\newblock \doi{10.1038/s41586-023-06728-8}.

\bibitem[Kucera et~al.(2022)Kucera, Togninalli, and Meng-Papaxanthos]{kucera2022ProteoGAN}
Kucera, T., Togninalli, M., and Meng-Papaxanthos, L.
\newblock Conditional generative modeling for de novo protein design with hierarchical functions.
\newblock \emph{Bioinformatics}, 38\penalty0 (13):\penalty0 3454--3461, 2022.

\bibitem[Lin et~al.(2023)Lin, Akin, Rao, Hie, Zhu, Lu, Smetanin, Verkuil, Kabeli, Shmueli, et~al.]{lin2023ESM2}
Lin, Z., Akin, H., Rao, R., Hie, B., Zhu, Z., Lu, W., Smetanin, N., Verkuil, R., Kabeli, O., Shmueli, Y., et~al.
\newblock Evolutionary-scale prediction of atomic-level protein structure with a language model.
\newblock \emph{Science}, 379\penalty0 (6637):\penalty0 1123--1130, 2023.

\bibitem[Ni et~al.(2023)Ni, Kaplan, and Buehler]{ni2023ProteinDiffusionGenerator}
Ni, B., Kaplan, D.~L., and Buehler, M.~J.
\newblock Generative design of de novo proteins based on secondary-structure constraints using an attention-based diffusion model.
\newblock \emph{Chem}, 9\penalty0 (7):\penalty0 1828--1849, 2023.

\bibitem[Nijkamp et~al.(2023)Nijkamp, Ruffolo, Weinstein, Naik, and Madani]{nijkamp2023progen2}
Nijkamp, E., Ruffolo, J.~A., Weinstein, E.~N., Naik, N., and Madani, A.
\newblock Progen2: exploring the boundaries of protein language models.
\newblock \emph{Cell systems}, 14\penalty0 (11):\penalty0 968--978, 2023.

\bibitem[Radford et~al.(2019)Radford, Wu, Child, Luan, Amodei, and Sutskever]{radford2019gpt2}
Radford, A., Wu, J., Child, R., Luan, D., Amodei, D., and Sutskever, I.
\newblock Language models are unsupervised multitask learners.
\newblock 2019.

\bibitem[Raffel et~al.(2020)Raffel, Shazeer, Roberts, Lee, Narang, Matena, Zhou, Li, and Liu]{2020t5}
Raffel, C., Shazeer, N., Roberts, A., Lee, K., Narang, S., Matena, M., Zhou, Y., Li, W., and Liu, P.~J.
\newblock Exploring the limits of transfer learning with a unified text-to-text transformer.
\newblock \emph{JMLR}, 21\penalty0 (140):\penalty0 1--67, 2020.

\bibitem[Rao et~al.(2021)Rao, Liu, Verkuil, Meier, Canny, Abbeel, Sercu, and Rives]{rao2021ESM_msa}
Rao, R.~M., Liu, J., Verkuil, R., Meier, J., Canny, J., Abbeel, P., Sercu, T., and Rives, A.
\newblock Msa transformer.
\newblock In \emph{ICML}. PMLR, 2021.

\bibitem[Repecka et~al.(2021)Repecka, Jauniskis, Karpus, Rembeza, Rokaitis, Zrimec, Poviloniene, Laurynenas, Viknander, Abuajwa, et~al.]{repecka2021proteinGAN}
Repecka, D., Jauniskis, V., Karpus, L., Rembeza, E., Rokaitis, I., Zrimec, J., Poviloniene, S., Laurynenas, A., Viknander, S., Abuajwa, W., et~al.
\newblock Expanding functional protein sequence spaces using generative adversarial networks.
\newblock \emph{Nature Machine Intelligence}, 3\penalty0 (4):\penalty0 324--333, 2021.

\bibitem[Rives et~al.(2021)Rives, Meier, Sercu, Goyal, Lin, Liu, Guo, Ott, Zitnick, Ma, et~al.]{rives2021ESM-1b}
Rives, A., Meier, J., Sercu, T., Goyal, S., Lin, Z., Liu, J., Guo, D., Ott, M., Zitnick, C.~L., Ma, J., et~al.
\newblock Biological structure and function emerge from scaling unsupervised learning to 250 million protein sequences.
\newblock \emph{Proceedings of the National Academy of Sciences}, 118\penalty0 (15):\penalty0 e2016239118, 2021.

\bibitem[Satorras et~al.(2021)Satorras, Hoogeboom, and Welling]{satorras2021n}
Satorras, V.~G., Hoogeboom, E., and Welling, M.
\newblock E(n) equivariant graph neural networks.
\newblock In \emph{ICML}, pp.\  9323--9332. PMLR, 2021.

\bibitem[Sillitoe et~al.(2021)Sillitoe, Bordin, Dawson, Waman, Ashford, Scholes, Pang, Woodridge, Rauer, Sen, et~al.]{sillitoe2021cath}
Sillitoe, I., Bordin, N., Dawson, N., Waman, V.~P., Ashford, P., Scholes, H.~M., Pang, C.~S., Woodridge, L., Rauer, C., Sen, N., et~al.
\newblock Cath: increased structural coverage of functional space.
\newblock \emph{NAR}, 49\penalty0 (D1):\penalty0 D266--D273, 2021.

\bibitem[Su et~al.(2024)Su, Ahmed, Lu, Pan, Bo, and Liu]{su2024RoPE}
Su, J., Ahmed, M., Lu, Y., Pan, S., Bo, W., and Liu, Y.
\newblock Roformer: Enhanced transformer with rotary position embedding.
\newblock \emph{Neurocomputing}, 568:\penalty0 127063, 2024.

\bibitem[Tan et~al.(2023)Tan, Zhou, Zheng, Fan, and Hong]{tan2023protssn}
Tan, Y., Zhou, B., Zheng, L., Fan, G., and Hong, L.
\newblock Semantical and topological protein encoding toward enhanced bioactivity and thermostability.
\newblock \emph{bioRxiv}, pp.\  2023--12, 2023.

\bibitem[Touw et~al.(2015)Touw, Baakman, Black, Te~Beek, Krieger, Joosten, and Vriend]{touw2015dssp_2}
Touw, W.~G., Baakman, C., Black, J., Te~Beek, T.~A., Krieger, E., Joosten, R.~P., and Vriend, G.
\newblock A series of pdb-related databanks for everyday needs.
\newblock \emph{NAR}, 43\penalty0 (D1):\penalty0 D364--D368, 2015.

\bibitem[Truong~Jr \& Bepler(2024)Truong~Jr and Bepler]{poet_neurips2023}
Truong~Jr, T. and Bepler, T.
\newblock Poet: A generative model of protein families as sequences-of-sequences.
\newblock \emph{NeurIPS}, 36, 2024.

\bibitem[Van~Kempen et~al.(2024)Van~Kempen, Kim, Tumescheit, Mirdita, Lee, Gilchrist, S{\"o}ding, and Steinegger]{van2024Foldseek}
Van~Kempen, M., Kim, S.~S., Tumescheit, C., Mirdita, M., Lee, J., Gilchrist, C.~L., S{\"o}ding, J., and Steinegger, M.
\newblock Fast and accurate protein structure search with foldseek.
\newblock \emph{Nature Biotechnology}, 42\penalty0 (2):\penalty0 243--246, 2024.

\bibitem[Vaswani et~al.(2017)Vaswani, Shazeer, Parmar, Uszkoreit, Jones, Gomez, Kaiser, and Polosukhin]{vaswani2017attention}
Vaswani, A., Shazeer, N., Parmar, N., Uszkoreit, J., Jones, L., Gomez, A.~N., Kaiser, {\L}., and Polosukhin, I.
\newblock Attention is all you need.
\newblock \emph{NeurIPS}, 30, 2017.

\bibitem[Watson et~al.(2023)Watson, Juergens, Bennett, Trippe, Yim, Eisenach, Ahern, Borst, Ragotte, Milles, et~al.]{watson2023RFDiffusion}
Watson, J.~L., Juergens, D., Bennett, N.~R., Trippe, B.~L., Yim, J., Eisenach, H.~E., Ahern, W., Borst, A.~J., Ragotte, R.~J., Milles, L.~F., et~al.
\newblock De novo design of protein structure and function with rfdiffusion.
\newblock \emph{Nature}, 620\penalty0 (7976):\penalty0 1089--1100, 2023.

\bibitem[Xie et~al.(2023)Xie, Valiente, and Kim]{xie2023helixgan}
Xie, X., Valiente, P.~A., and Kim, P.~M.
\newblock Helixgan a deep-learning methodology for conditional de novo design of $\alpha$-helix structures.
\newblock \emph{Bioinformatics}, 39\penalty0 (1):\penalty0 btad036, 2023.

\bibitem[Yang et~al.(2023)Yang, Zanichelli, and Yeh]{yang2023MIFST}
Yang, K.~K., Zanichelli, N., and Yeh, H.
\newblock Masked inverse folding with sequence transfer for protein representation learning.
\newblock \emph{Protein Engineering, Design and Selection}, 36:\penalty0 gzad015, 2023.

\bibitem[Yeh et~al.(2023)Yeh, Norn, Kipnis, Tischer, Pellock, Evans, Ma, Lee, Zhang, Anishchenko, et~al.]{yeh2023novo}
Yeh, A. H.-W., Norn, C., Kipnis, Y., Tischer, D., Pellock, S.~J., Evans, D., Ma, P., Lee, G.~R., Zhang, J.~Z., Anishchenko, I., et~al.
\newblock De novo design of luciferases using deep learning.
\newblock \emph{Nature}, 614\penalty0 (7949):\penalty0 774--780, 2023.

\bibitem[Yi et~al.(2024)Yi, Zhou, Shen, Li{\`o}, and Wang]{yi2024GRADE_IF}
Yi, K., Zhou, B., Shen, Y., Li{\`o}, P., and Wang, Y.
\newblock Graph denoising diffusion for inverse protein folding.
\newblock \emph{NeurIPS}, 36, 2024.

\bibitem[Yim et~al.(2024)Yim, Campbell, Mathieu, Foong, Gastegger, Jim{\'e}nez-Luna, Lewis, Satorras, Veeling, No{\'e}, et~al.]{yim2024SE3flow_match}
Yim, J., Campbell, A., Mathieu, E., Foong, A.~Y., Gastegger, M., Jim{\'e}nez-Luna, J., Lewis, S., Satorras, V.~G., Veeling, B.~S., No{\'e}, F., et~al.
\newblock Improved motif-scaffolding with {SE(3)} flow matching.
\newblock \emph{arXiv:2401.04082}, 2024.

\bibitem[Zhang \& Skolnick(2005)Zhang and Skolnick]{zhang2005tm-align}
Zhang, Y. and Skolnick, J.
\newblock Tm-align: a protein structure alignment algorithm based on the tm-score.
\newblock \emph{NAR}, 33\penalty0 (7):\penalty0 2302--2309, 2005.

\bibitem[Zheng et~al.(2022)Zheng, Lu, Zan, Li, Liu, Liu, Huang, Liu, et~al.]{zheng2022loosely}
Zheng, L., Lu, H., Zan, B., Li, S., Liu, H., Liu, Z., Huang, J., Liu, Y., et~al.
\newblock Loosely-packed dynamical structures with partially-melted surface being the key for thermophilic argonaute proteins achieving high dna-cleavage activity.
\newblock \emph{NAR}, 50\penalty0 (13):\penalty0 7529--7544, 2022.

\bibitem[Zhou et~al.(2023)Zhou, Zheng, Wu, Yi, Zhong, Lio, and Hong]{zhou2023CPDiffusion}
Zhou, B., Zheng, L., Wu, B., Yi, K., Zhong, B., Lio, P., and Hong, L.
\newblock Conditional protein denoising diffusion generates programmable endonucleases.
\newblock \emph{bioRxiv}, pp.\  2023--08, 2023.

\end{thebibliography}
\bibliographystyle{icml2024}




\end{document}